\documentclass[11pt,a4paper]{article}

\usepackage[hyperref]{acl2020}
\usepackage{times}
\usepackage{latexsym}
\usepackage[utf8]{inputenc} %
\usepackage{comment}
\usepackage{url}           %
\usepackage{booktabs}       %
\usepackage{amsfonts}       %
\usepackage{nicefrac}       %
\usepackage{microtype}      %
\usepackage{graphicx}       %
\usepackage{multirow}
\usepackage{tabularx} 
\usepackage{enumitem}
\newlist{compactitem}{itemize}{3} %
\setlist[compactitem]{label=\textbullet, leftmargin=1em, nosep}
\usepackage{dblfloatfix}
\usepackage{color}
\usepackage{dirtytalk}
\usepackage{ntheorem}

\newcommand{\xhdr}[1]{\vspace{1mm}\noindent{{\bf #1.}}}

\theoremseparator{:}
\newtheorem{hyp}{Hypothesis}
\aclfinalcopy %
\def\aclpaperid{3129} %

\usepackage{amsmath,amsfonts,bm}

\def\eqref#1{equation~\ref{#1}}

\def\1{\bm{1}}

\def\rva{{\mathbf{a}}}

\DeclareMathAlphabet{\mathsfit}{\encodingdefault}{\sfdefault}{m}{sl}
\SetMathAlphabet{\mathsfit}{bold}{\encodingdefault}{\sfdefault}{bx}{n}

\newcommand{\astatecap}{{G}}
\newcommand{\astate}{{\intstate}}
\newcommand{\castate}{{\astate_{t}}}

\newcommand{\Ucal}{{\mathcal U}}

\newcommand{\intaction}{{\tilde{a}}}
\newcommand{\intstate}{{\tilde{s}}}
\newcommand{\astateset}{{\mathcal \astatecap}}
\newcommand{\astatecapstar}{{\astateset^{*}}}
\newcommand{\recog}{{O^{-1}}}
\newcommand{\pol}{{\pi}}
\newcommand{\Temu}{{\tilde{T}}}
\newcommand{\dnote}[1]{\textcolor{orange}{$\ll$\textsf{#1 --Dafna}$\gg$}}
\newcommand{\rnote}[1]{\textcolor{green}{$\ll$\textsf{#1 --Ronen}$\gg$}}

\newcommand{\tnote}[1]{\textcolor{cyan}{$\ll$\textsf{#1 --Tom}$\gg$}}
\newcommand{\remove}[1]{}
\newcommand\tab[1][0.7cm]{\hspace*{#1}}

\newcommand{\affa}{{$^{\dagger}$}}
\newcommand{\affb}{{$^{\ddagger}$}}
\newcommand{\affc}{{$^{\star}$}}
\newcommand{\affd}{{$^{\ast}$}}

\title{Language (Re)modelling: Towards Embodied Language Understanding}

\author{Ronen Tamari\affa \tab Chen Shani\affa \tab Tom Hope\affc\affd \\ \quad {\bf Miriam R. L. Petruck}\affb \tab  {\bf Omri Abend}\affa \tab {\bf Dafna Shahaf}\affa \\
  \affa The Hebrew University of Jerusalem \\
  \affc Allen Institute for Artificial Intelligence \\
  \affd Paul G. Allen School of Computer Science \& Engineering, University of Washington \\
  \affb International Computer Science Institute, Berkeley, CA \\
  \texttt{\{ronent,chenxshani,oabend,dshahaf\}@cs.huji.ac.il} \\
  \texttt{tomh@allenai.org} \tab \texttt{miriamp@icsi.berkeley.edu}
}
 
\date{}

\begin{document}

\maketitle

\begin{abstract}
While natural language understanding (NLU) is advancing rapidly, today's technology differs from human-like language understanding in fundamental ways, notably in its inferior efficiency, interpretability, and generalization. This work proposes an approach to representation and learning based on the tenets of embodied cognitive linguistics (ECL). According to ECL, natural language is inherently executable (like programming languages), driven by mental simulation and metaphoric mappings over hierarchical compositions of structures and schemata learned through embodied interaction. This position paper argues that the use of grounding by metaphoric inference and simulation will greatly benefit NLU systems,  and proposes a system architecture along with a roadmap towards realizing this vision.

\end{abstract}

\section{Introduction}

 \begin{quotation}
\noindent ``Not those speaking the same language, but those sharing the same feeling understand each other.'' \textit{-- Jalal ad-Din Rumi}
 \end{quotation}

While current NLU systems ``speak'' human language by learning strong statistical models, they do not possess anything like the rich mental representations that people utilize for language understanding. Indeed, despite the tremendous progress in NLU, recent work shows that today's state-of-the-art (SOTA) systems differ from human-like language understanding in crucial ways, in particular in their generalization, grounding, reasoning, and explainability capabilities~\citep{glockner-etal-2018-breaking,mccoy2019berts,mccoy-etal-2019-right,Nie2019, yogatama2019learning,lake2019human}.

Question-answering (QA) is currently one of the predominant methods of training deep-learning models for general, open-domain language understanding~\citep{gardner2019question}. While QA is a versatile, broadly-applicable framework, recent studies have shown it to be fraught with pitfalls~\citep{Gardner2019, Mudrakarta2018}. A recent workshop on QA for reading comprehension suggested that ``There is growing realization that the traditional supervised learning paradigm is broken [...] %
-- we’re fitting artifacts''~\citep{mrqa_pres2019}.

In many respects, the problems of NLU mirror those of artificial intelligence (AI) research in general. \citeauthor{Lake2017}'s (\citeyear{Lake2017}) seminal work identified a significant common factor at the root of problems in general AI. The current deep-learning paradigm is a statistical pattern-recognition approach predominantly applied to relatively narrow \emph{task-specific prediction}. In contrast, human cognition supports a  wide range of inferences (planning, action, explaining, etc.), hinting at a view of intelligence focused on \emph{model-building}, specifically, mental models: rich, structured, manipulable, and explainable representations %
useful for performing in dynamic, uncertain environments. This distinction motivates the quest for a new {\bf cognitively-inspired model-building} learning paradigm for general AI, which has inspired fruitful subsequent research and discussion (e.g., \citet{Lake2017b}).

The observation that NLU and general AI share a common central problem (task-specific prediction-based learning), and the growing realization that deeper text understanding requires building mental models~\citep{Gardner2019,forbes2019neural}, motivate the search for an \emph{NLU analog} of the cognitively-inspired {model building paradigm}.

Amid recent position papers highlighting significant differences between human language understanding and current NLU systems~\citep{McClelland2019,bisk2020experience}, here we take a more focused look at mental models; challenges arising due to their embodied nature, their importance in general NLU, and how we might begin integrating them into current approaches.

Mainstream NLU work, be it entirely distributional, such as BERT \citep{devlin-etal-2019-bert}, or also involving symbolic knowledge representation \citep{liu2019k, bosselut-etal-2019-comet}, seldom addresses mental models %
directly. Crucially, such approaches lack the interactive \emph{worlds} within which mental models\footnote{Typically, mental models are construed as ``world simulators''; see \S\ref{sec:emb-cog-paradigm}.} are learned jointly through language and embodied action. The most closely related lines of work to the present proposal are grounded approaches, which feature worlds in the form of interactive environments, and address mapping text to programs (executable semantic parses) \citep[e.g.,][]{Mooney2008AAAI,artzi-zettlemoyer-2013-weakly,gauthier2016paradigm,liang2016learning,kiela2016virtual,Boisvert19}. 
However, while well-aligned with a model-building paradigm, typically such approaches have been limited to short or synthetic literal language and narrow domains assuming predefined environments.
Embodied approaches to \emph{general} NLU, as advocated here, are few and far between. Mostly, examples fall under the construction grammar framework~\citep{steels-de-beule-2006-brief,bergen2005simulation}. However, despite their intellectual merit, they were not operationalized to scale readily for mainstream applications (see \S \ref{sec:emb-cog-paradigm}).

This position paper argues that executable semantic parsing and grounded approaches to NLU constitute a first step in a much larger program, whose outline is set forth, for \emph{general} language understanding through {\bf embodied cognitive linguistics (ECL)}. Following much cognitive science research (see \S\ref{sec:emb-cog-paradigm}, \S\ref{sec:ecl-link}), this paper posits that (1) {\bf execution or simulation} {is a central part of semantics}, essential for addressing some of the persistent difficulties in text understanding, and (2) \textbf{metaphoric inference capabilities} are central to {knowledge representation}, and facilitate grounded understanding of general language. Importantly, capacities for both simulation and metaphor are emergent, borne of embodied interaction within an external world.

Our contributions are: we analyze inherent limitations of SOTA statistical language models applied to NLU and propose a framework to address these limitations. The novelty of this approach stems from bringing together ideas from the cognitive science literature, the wider AI community, and NLU. This framework constitutes a path to generalize current execution-based methods towards more general language understanding. 

This paper proposes a system architecture and a roadmap towards implementing the vision outlined here, suggesting preliminary directions for future work (learned world models, incorporating interaction into datasets). %
 We believe that this framework will facilitate consolidation with multiple related lines of research across the different communities, particularly embodied AI and NLU~\citep{Luketina2011}.

\section{Challenges for Current NLU Systems}
\label{sec:challenges}
This section presents concrete example problems demonstrating inherent limitations in SOTA NLU.\label{challenges} %

\subsection{Open-domain Literal Language Simulation}
\label{ssec:ootb-reason}

\begin{figure}[t!]
\centering
\includegraphics[width=0.98\linewidth]{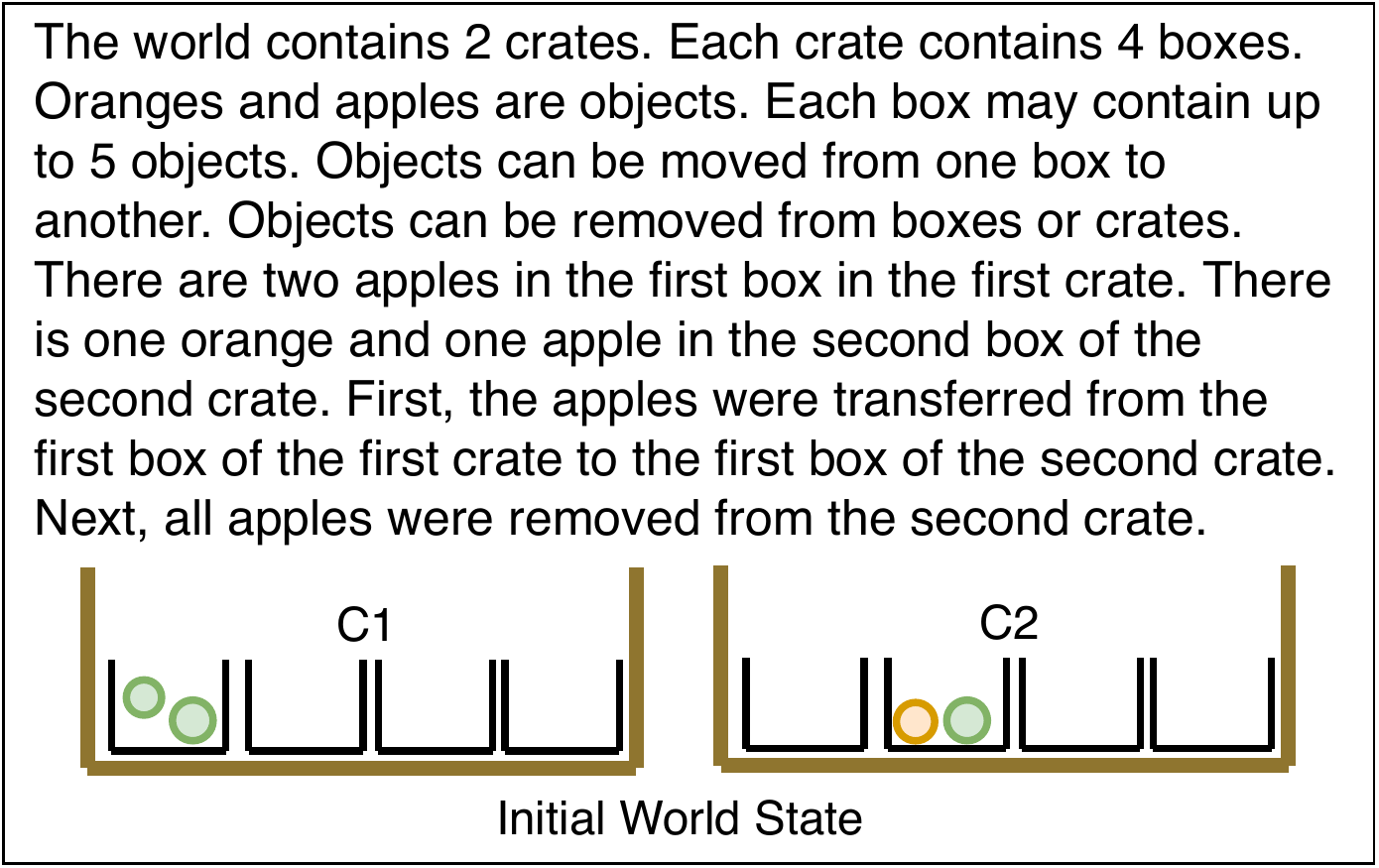}
\caption{\label{fig:scone-in-a-box} Open-domain challenge -- a world with boxes, crates and objects. %
}
\end{figure}

Fig.~\ref{fig:scone-in-a-box} includes a short story about a world with crates, boxes, and objects inside them. %
It is a short and simple narrative, far from capturing the full-blown complexity of natural language. Following \citet{Gardner2019}, we assume that a system understands the story if it can correctly answer arbitrary questions about it. To do so requires basic commonsense and mathematical reasoning, referent grounding, tracking events, handling declarative knowledge, and more.  

The task is similar to narrative comprehension tasks in datasets such as bAbI~\citep{bordes2015artificial} and SCONE~\citep{long-etal-2016-simpler}, and could be solved given large amounts of annotated training data. But, the goal here is different, specifically, to
develop models that, like humans, can understand such language \textbf{on-the-fly} (like zero-shot learning).

\xhdr{QA approaches}
Current QA systems, used in an off-the-shelf manner, do not generalize well to tasks on which they have not been trained; NLU models are known to be brittle even to slight changes in style and vocabulary~\citep{gardner2020evaluating,Keysers2020}. The closest QA setting is the DROP challenge~\citep{dua-etal-2019-drop}, requiring reading comprehension and basic numerical reasoning over paragraphs. As a simple sanity check, we tested a near-SOTA model and baseline\footnote{\citet{Efrat2019TagbasedME} and \citet{dua-etal-2019-drop}, respectively.} on this example, asking questions about the initial and final state. The models were notably better answering questions about the initial state than about the final state. This result is perhaps expected, as the answers to questions about the initial state are closer to the input text. Answering questions about later states is more challenging. A key missing component of these systems is the ability to \textbf{simulate} the effects of actions, especially commonsense effects (e.g., moving a container moves the elements in it).

\xhdr{Executable semantic parsing approaches}
The problem of Fig.~\ref{fig:scone-in-a-box} could also naturally be cast as an executable semantic parsing task. Similar tasks already exist, for example, the ``Alchemy'' sub-task of the SCONE dataset features beakers of chemicals that are mixed, poured, and drained. Executable approaches can leverage simulation to learn structured world models, but are limited by hard-coded, domain-specific executors; adding tasks requires substantial manual effort.

For humans, through largely subconscious metaphorical inference (related to transfer and meta-learning in general AI~\citep{Lake2017}), it is obvious that both SCONE and Fig.~\ref{fig:scone-in-a-box} share much the same structure. This similarity allows for effortless generalization, effectively re-purposing a relatively simple executor (for literal language) flexibly across many tasks. 

\begin{figure}[t!]
\includegraphics[width=0.46\textwidth]{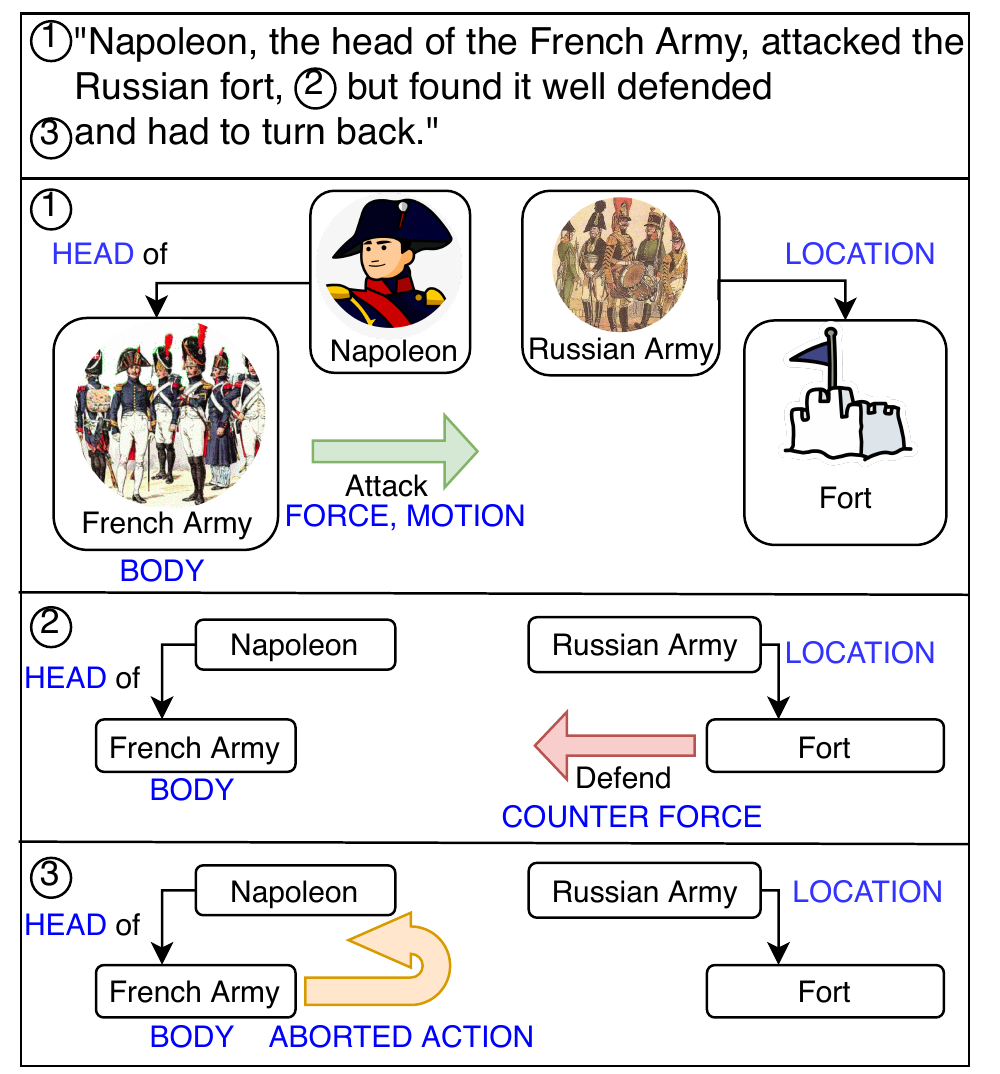}
\caption{\label{fig:metalang-challenge} Non-literal language challenge. To understand this sentence, humans rely on metaphoric inference over embodied concepts (in blue, also called schema; see \S\ref{sec:emb-cog-paradigm}). 
For example, here ``attack'' evokes a FORCE or MOTION schema, used to construct a mental model of the scene via mental simulation (\S\ref{sec:ecl-link}). %
} 
\end{figure}

\subsection{Non-literal Language}
\label{ssec:meta-reason}

The previous challenge involved literal language, amenable to symbolic execution. However, 
non-literal language is pervasive in everyday speech~\citep{lakoff1980metaphorical}. Consider the example in Fig.~\ref{fig:metalang-challenge}: the phrase ``head of the French Army'' is non-literal, implying that the army can be treated as a human body. The execution semantics of verbs like ``attacked'' and ``defend'' are also non-literal; they are highly contextual, requiring interpretation beyond word-sense disambiguation alone. %
``Russian hackers attacked the Pentagon networks'' or ``The senator attacked the media'' entail very different \emph{simulations}. This ambiguity is  challenging for non-neural (symbolic) simulation-based approaches. Humans compose a structured mental model from the language through schemata and mental simulation, as discussed in \S\ref{sec:emb-cog-paradigm},\S\ref{sec:ecl-link}.

To summarize, the limitations outlined above motivate the attempt to extend the capability of simulation to general linguistic inputs. Doing so would enable the construction of grounded, manipulable, and interpretable representations from text. Two desiderata follow from the challenges: (1) more flexible utilization of symbolic executors by exploiting shared (analogical) structures between texts (\S\ref{ssec:ootb-reason}), and (2) learned, neural executors for  non-literal language comprehension (\S\ref{ssec:meta-reason}).

\section{Embodied Cognitive Linguistics: A Model Building Paradigm}
\label{sec:emb-cog-paradigm}

Turning to cognitive science for inspiration, we focus on embodied cognitive linguistics (ECL), an important paradigm directly addressing both desiderata. This section presents a brief overview and key tenets of ECL, specifically the theoretical foundations ~\citet{lakoff1980metaphorical} and ~\citet{Feldman2004} developed. Most contemporary cognitive accounts of language incorporate concepts from ECL to some degree. A full review is out of scope of this work; see \citet{gardenfors2014geometry} and \S\ref{sec:ecl-link},\S\ref{sec:ecl-language-model} for discussion in the NLU context.

Early cognitive theories assumed a disembodied, symbolic representation of knowledge~\citep{lewis1976general,kintsch1978toward}, separate from the brain's modal systems (vision, motor control, etc.). In contrast, the embodied cognition (EC) view, based on widespread empirical findings, focuses on the role of the body in cognition. In this view, knowledge is stored using multi-modal representations (mental imagery, memories, etc.) that arise from embodied experience and action in the world~\citep{Barsalou2008, proffitt2006embodied}. 
ECL postulates that linguistic representations and other, higher-level cognitive functions are deeply grounded in neural modal systems~\citep{lakoff1980metaphorical,Barsalou2008}. This view is compelling, as it addresses the grounding problem~\citep{HARNAD1990335} by linking between high-level symbolic constituents of mental representations and experience or action in the physical world~\citep{varela2017embodied}. Note that embodiment is far from an end-all for language comprehension: for example, social and cultural aspects too are crucial~\citep{Arbib2014}. Still, ECL laid important conceptual foundations also underlying subsequent accounts:

\begin{compactitem}
\item \textbf{Embodied schemata:} Pre-linguistic structures formed from bodily interactions and recurring experience, such as CONTAINMENT, PART-WHOLE, FORCE, MOVEMENT~\citep{Langacker19871991, talmy1985lexicalization,talmy1983language}.

\item \textbf{Metaphoric inference\footnote{Also called analogical reasoning, we use ``metaphorical'' and ``analogical'' interchangeably.}:} The process by which new information may be inferred via structural similarities to a better-understood instantiated system~\citep{lakoff1980metaphorical,Gallese2005,Day2007}. For example, ``I have an example {IN} mind'' suggests that %
the abstract concept \textit{mind} is mapped to the more concrete domain of \textit{containers}.

\item \textbf{Mental simulation:} The reenactment of perceptual, motor, and introspective states acquired during experience with the world, body, and mind. In EC, diverse simulation mechanisms (also called mental or forward models~\citep{SchemataThoughtPDP1986,Grush2004}) support a wide spectrum of cognitive activities, including language and decision making~\citep{Barsalou2008}.
\end{compactitem}

We believe that ECL is a useful paradigm for addressing the challenges of \S\ref{sec:challenges}, as it articulates the role of analogy and mental simulation in NLU. The following two ECL hypotheses summarize them~\citep{lakoff1980metaphorical,Feldman2004}:

\begin{hyp}[Simulation] \label{hyp:sim}
Humans understand the meaning of language by mentally simulating its content. Language in context evokes a simulation structured by embodied schemata and  metaphoric mappings, utilizing the same neural structures for action and perception in the environment. Understanding involves inferring and running the best fitting simulation.
\end{hyp}

\begin{hyp}[Metaphoric Representation] \label{hyp:metarep}
Human concepts are expressible through hierarchical, compositional, metaphoric mappings over a limited vocabulary of embodied schema. Abstract concepts are expressed using more literal concepts. 
\end{hyp}

\xhdr{Early ECL Implementations}
Early attempts to implement ECL in actual language understanding systems were founded on \citet{narayanan1997knowledge}'s x-schema simulation framework and Embodied Construction Grammar~\citep{bergen2005simulation}. While notable for approaching challenging problems involving mental simulation, and complex, metaphoric language, early implementation efforts were not operationalized to scale to mainstream applications~\citep{lakoff2010toward}. These works also focused on a particular type of simulation (sensorimotor), understood as only one mechanism of many used in language understanding~\citep{Stolk2016}.

FrameNet~\citep{Ruppenhofer16} and MetaNet~\citep{david2014building} are closely related projects in that each provides an extensive collection of schemata used in everyday and metaphoric language comprehension, respectively, via the concept of a semantic frame~\citep{fillmore1985frames}. However, neither incorporates simulation semantics, as needed for a full realization of the ECL vision~\citep{chang-etal-2002-putting}.

\section{Linking ECL to NLU and Embodied AI Research}
\label{sec:ecl-link}

\begin{table*}[t!]
\small
\begin{tabularx}{\textwidth} { 
   >{\centering\arraybackslash}X 
  | >{\centering\arraybackslash}X 
  | >{\centering\arraybackslash}X
  | >{\centering\arraybackslash}X   }
 
 \textbf{Concept}    &  \textbf{Symbolic}         & \textbf{ECL} & \textbf{Embodied AI}  \\ 
 \toprule
  \textbf{Primitives} & Basic data structures, operators, variables...  &   {Schemata:}  MOVE, CONTAINER, PART-WHOLE... & \multirow{2}{*}{\begin{tabular}[c]{@{}c@{}}\\[1em]Deep neural world \& \\ action representations \\ (learned through interaction)\end{tabular}} \\ \cline{1-3}
 \textbf{Knowledge Organization} & a) Composition, inheritence \newline b) Libraries 
 &  a) Hierarchical, compositional metaphoric mappings \newline b) Compiled Knowledge & 
 \\
\hline
 \textbf{Executable Unit} & Instruction & {Semantic parse} & $\intaction$\\
\hline
\\[-1em]
 \textbf{Execution Trace} & Intermediate program states & {Mental models} & $\Temu\left(\intstate,\intaction\right)$  \\
 \hline
 \\[-1em]
 \textbf{Simulation} & Executor & {Emulator}\affa & $\Temu$  \\
\hline
\\[-1em]
 \textbf{Semantic parsing / grounding } & Parser to executable symbolic program & {Parser to executable neural program} & $O^{-1},\pol$  \\
\bottomrule
\end{tabularx}
\caption{\label{tab:programming-analogy}  \textit{Natural language as a neural programming language} conceptualization, with correspondence between symbolic programming, ECL, and embodied AI, using standard POMDP notation. Tilde notation refers to internal counterparts of $T,s,a$ used in mental simulation. \\ \affa Also called mental simulation~\citep{bergen2005simulation}, we adopt \textit{emulator} \citep{Glenberg2008} to conform with contemporary cognitive science accounts.}
\end{table*}
 We propose a unifying view of ECL, bringing it closer to contemporary cognitive science and deep learning approaches. This section presents notations and motivating intuitions, further developing the computational framework in \S\ref{sec:ecl-language-model},\S\ref{sec:emb-arch}.
The proposal centers around the view of natural language as a kind of \emph{neural programming language}~\citep{Lupyan2016}, or higher-level cognitive control system for systematically querying and inducing changes in the mental and physical states of recipients~\citep{elman2004alternative,Stolk2016,Borghi2018}. This approach builds on the ECL hypotheses and suggests a broader view of mental simulation, one that is readily amenable to the same computational formulation as current embodied AI and executable semantic parsing approaches. 
 
  \xhdr{Preliminaries} At the core of embodied approaches is the Partially Observable Markov Decision Process \citep[POMDP;][]{kaelbling1998planning}. It governs the relations between states ($s$), actions ($a$), observations ($o$), and rewards ($r$). Of particular interest are the recognition $\recog:O\to S$, policy $\pi:S\to A$, and transition  $T:S\times A\to S$ functions. Focusing on mental simulation rather than actual external action, we assume a degree of equivalence between external and internal representations~\citep{SchemataThoughtPDP1986,Hamrick2019}. We consider internal mental states and actions ($\intstate, \intaction$), effecting change to mental models via a learned \textbf{neural emulator} $\Temu$~\citep{Grush2004}. Finally, language is considered a form of action~\citep{Glenberg2008} via external and internal utterances (i.e., semantic parses).

\xhdr{Connecting symbolic \& embodied language understanding}
Table \ref{tab:programming-analogy} presents a structured version of the neural programming language conceptualization. Importantly, this view highlights the important commonalities and differences between ECL and both \textbf{symbolic programming languages}, as well as \textbf{embodied neural mechanisms}, for perception and action. We illustrate these relations more explicitly through a comparison between ECL and executable semantic parsing (Table \ref{tab:programming-analogy}, bottom).

 \xhdr{Executable semantic parsing} Involves parsing a novel linguistic input $o$ into a symbolic program $\rva$, whose execution\footnote{Slightly abusing notation, we apply $T$ iteratively on a sequence of actions $\rva=\left(a_{0},...,a_{L-1}\right)$.} yields a desired goal state: $T\left(O^{-1}\left(o\right),\rva\right)=s^{*}$. Executable semantic parsing focuses on action in an external, symbolic environment $T$, and typically doesn't address $\Temu$, e.g., mapping a natural language question $o$ directly to an executable query $\rva$ on an SQL engine $T$.

\xhdr{ECL semantic parsing} Shares the same structure as executable semantic parsing, with the important distinction that simulation is enacted via internal neural representations: $\Temu\left(O^{-1}\left(o\right),\tilde{\rva}\right)=\intstate^{*}$. The fully neural formulation enables grounded understanding of non-literal language, demonstrated here for the Fig. \ref{fig:metalang-challenge} example. \textbf{Metaphoric inference (hyp. \ref{hyp:metarep})} facilitates parsing a novel linguistic input $o$ into \textbf{internal, structured, neural} state representations $\intstate$, $\intaction$. Accordingly, the utterance $u$=``Napoleon, the head of the French Army'' might be parsed to an internal state $\intstate$ composed of a PART-WHOLE schema as shown in the figure. The phrase ``attacked the Russian fort'' could be grounded to a parse $\intaction$ driving simulation over MOTION and FORCE schemata. The requirement that $\intstate$ and $\intaction$ should afford \textbf{mental simulation (hyp. \ref{hyp:sim})} by the neural world emulator $\Temu$ marks an important difference from current neural word embeddings, one that contributes to deeper language understanding; in the resulting mental model $\Temu\left(\intstate,\intaction\right)$, Napoleon and the French Army likely moved \emph{together} due to the PART-WHOLE relation between them. This inference is non-trivial since it requires implicit knowledge (heads and bodies often move together). Indeed, a SOTA NLI model\footnote{We use \citet{liu2019roberta} with \url{https://demo.allennlp.org/textual-entailment/}.} considers it ``very likely'' that the Fig. \ref{fig:metalang-challenge} sentence contradicts the entailment that ``The French Army moved towards the fort but did not enter it.'' To summarize: 
\begin{compactitem}
 \item Executable semantic parsing approaches address grounding \textbf{literal} language to \textbf{symbolic} primitives, whereas metaphoric inference suggests a mechanism for grounding \textbf{general} language using \textbf{neural} primitives (schemata).
 \item Executable semantic parsing approaches utilize \textbf{hard-coded, external symbolic executors}, whereas ECL highlights the role of \textbf{learned neural world emulators}, as in current embodied research AI efforts (see \S\ref{ssec:phase-two}).
\end{compactitem}

\section{Proposal for an Embodied Language Understanding Model}

\label{sec:ecl-language-model}

 Formalizing the view characterized above suggests a novel computational model of language understanding. While current statistical models focus on the \emph{linguistic signal}, research shows that most of the relevant information required for understanding a linguistic message {is not present in the words}~\citep{Stolk2016,David2016}. Accordingly, the ECL view suggests shifting the focus to the \emph{mental models} that communicators use, and the neural mechanisms used to construct them, e.g., mental simulation.
 
 What follows here adapts a relevant cognitive-inspired framework from general AI to the present NLU setting (\S\ref{ssec:comm-model}), and discusses computational challenges (\S\ref{ssec:emb-challenges}). Note that similar insights have been applied to multi-agent communication problems~\citep{Andreas2017}, but their application to general NLU has been limited.

\subsection{Formal Framework}
\label{ssec:comm-model}

\begin{figure*}[t!]
\centering
\includegraphics[width=0.64\textwidth]{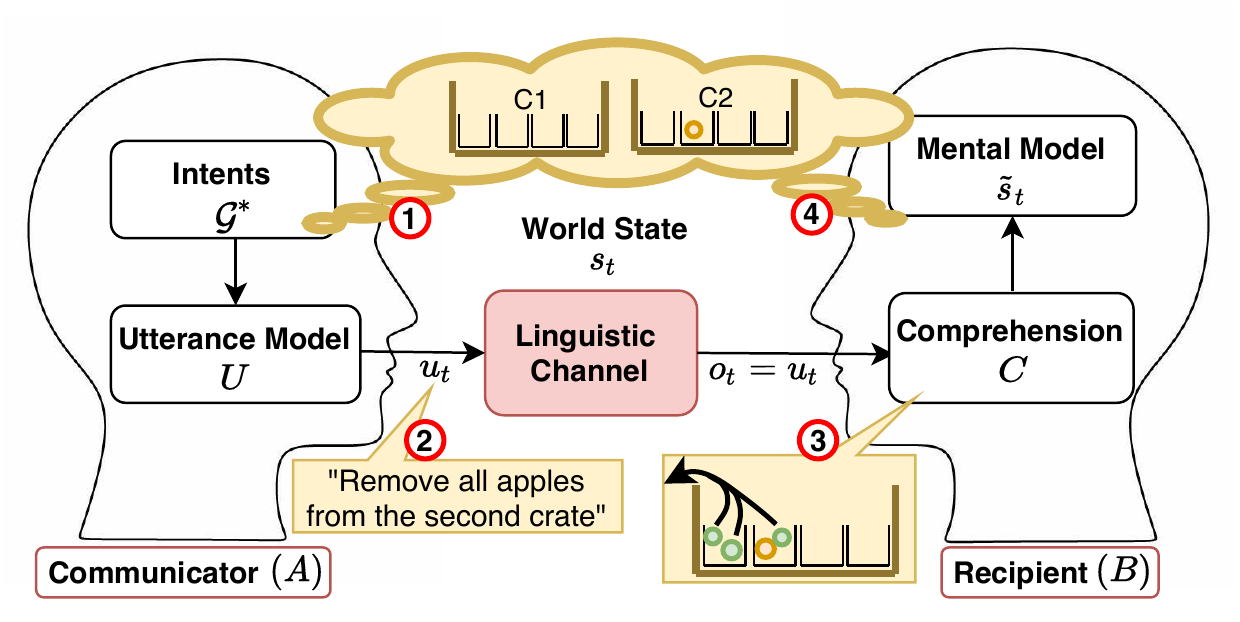}
\caption{\label{fig:comm_model} Schema of linguistic communication framework. Communicator's intent (1) is a high dimensional mental state, i.e., remove apples from the second crate. The low capacity of the linguistic channel (2) leaves the burden of understanding primarily on Communicator and Recipient (embodiment principle). The Recipient's goal is to understand (3), i.e., reconstruct the intent by integrating linguistic input, knowledge of the state of the world, and internal knowledge (memories, commonsense). Reconstruction results in a successful alignment (4).}
\end{figure*}

The recently introduced Consciousness Prior~\citep[CP;][]{Bengio2017} is a framework to represent the mental model of a single agent, through the notion of \emph{abstract state representations}.\footnote{For brevity we omit discussion of deriving abstract states from the full mental state, see \citet{Bengio2017} for details.} Here, an abstract state corresponds with $\astate$ (\S\ref{sec:ecl-link}), a low-dimensional, structured, interpretable state encoding, useful for planning, communication, and predicting upcoming observations~\citep{franccois2019combined}. One example is a dynamic knowledge graph embedding to represent a scene~\citep{Kipf2020Contrastive}.

We adapt CP to a two-player cooperative linguistic communication setting~\citep{Tomasello2008}. We assume a communicator ($A$) and recipient ($B$), as shown in Fig. \ref{fig:comm_model}. The computational problem of communicators is a ``meeting of minds''~\citep{gardenfors2014geometry}, or achieving some \textbf{alignment} of their \textbf{mental models} %
 \citep{rumelhart1991understanding,Stolk2016}: the communicator $A$ wishes to induce in $B$ some (possibly ordered) set of goal abstract states $\astatecapstar$.

We leave exploration of the communicator side to future work, and focus here on understanding. We assume that $A$ sequentially generates utterances $u_{t} \in \Ucal$ (we assume equivalence between utterances $u$ and observations $o$) using an utterance model~\citep{Bengio2017}. Analogously, $B$ uses a \textit{comprehension} model $C$ s.t., $\castate=C\left(\astate_{t-1},u_{t}\right)$. %
We assume that alignment is possible: there exists some sequence of utterances that will induce $\astatecapstar$. 

This framework is readily applicable to static text (reading comprehension).  For example, in Fig. \ref{fig:scone-in-a-box}, $\astatecapstar$ would be the sequence of desired states, and each sentence corresponds to an utterance  ($u_1=$``The world contains 2 crates.'',...).

\subsection{Computational challenges of embodiment}
\label{ssec:emb-challenges}
We can now more precisely characterize the challenges that the recipient faces. At the root of the problem is the \textit{embodiment principle}~\citep{Lawrence2017}: human internal representations and computation capacity, as represented by $\intstate$ and $\Temu$, respectively, are many orders of magnitude larger than their linguistic communication ``bandwidth''. We note that though $\castate$ is only a subspace of the full mental state, following \citet{Stolk2016,Bengio2017} we assume that it still holds that $\text{dim}\left(\castate\right)\gg\text{dim}\left(u_{t}\right)$.The embodiment principle dictates extreme economy in language use~\citep{grice1975logic}, and results in three major challenges:

\xhdr{Common ground (prior world knowledge)} Meaning cannot be spelled out in words but rather must be \textbf{evoked} in the listener~\citep{rumelhart1991understanding} by assuming and exploiting common ground~\citep{clark1989contributing,Tomasello2008}, i.e., shared structures of mental representations. In other words, to achieve some aligned goal state $g^{*}$, the communicators must rely heavily on pre-existing similarities in $\intstate$, $\intaction$, and $\Temu$. Developing computational versions of human world models ($\Temu$) is likely AI-complete or close, but useful middle ground may be attained by partial approximations.

\xhdr{Common ground (discourse)} In the context of discourse, new information must be accumulated efficiently to update the mental model~\citep{clark1989contributing,Stolk2016}. Consider ``Remove all apples from the second crate'' (Figure \ref{fig:scone-in-a-box}). Full comprehension is only possible in the context of a sufficiently accurate mental model. Using our previous notations, the comprehension of $u_t$ depends both on  the previous utterances $u_{1:\left(t-1\right)}$ and intermediate mental model $\intstate_{t-1}$.

\xhdr{Abstract vs. Literal Language}  Interpretation of literal language is relatively straightforward -- it is the language first acquired by children, directly related to the physical world. However, much of human language  
is more abstract, relying on %
\emph{metaphors} borne of embodiment.  The symbolic programming analog fails for utterances like ``these elections seem like a circus''. Symbolic programming languages cannot handle non-literal interpretations: \emph{how} are \texttt{elections} like a \texttt{circus}? This is related to \emph{selective analogical inference}~\citep{gentner2011computational}, closely related to ECL: not everything in the source domain (circus) is mapped to the target (elections). Humans easily perceive the salient metaphoric mappings (\texttt{clown}$\to$\texttt{candidate}), but this feat remains extremely complex for machines.

\section{Architecture Sketch}
\label{sec:emb-arch}

\begin{figure*}[t!]
\centering
\includegraphics[width=0.94\linewidth]{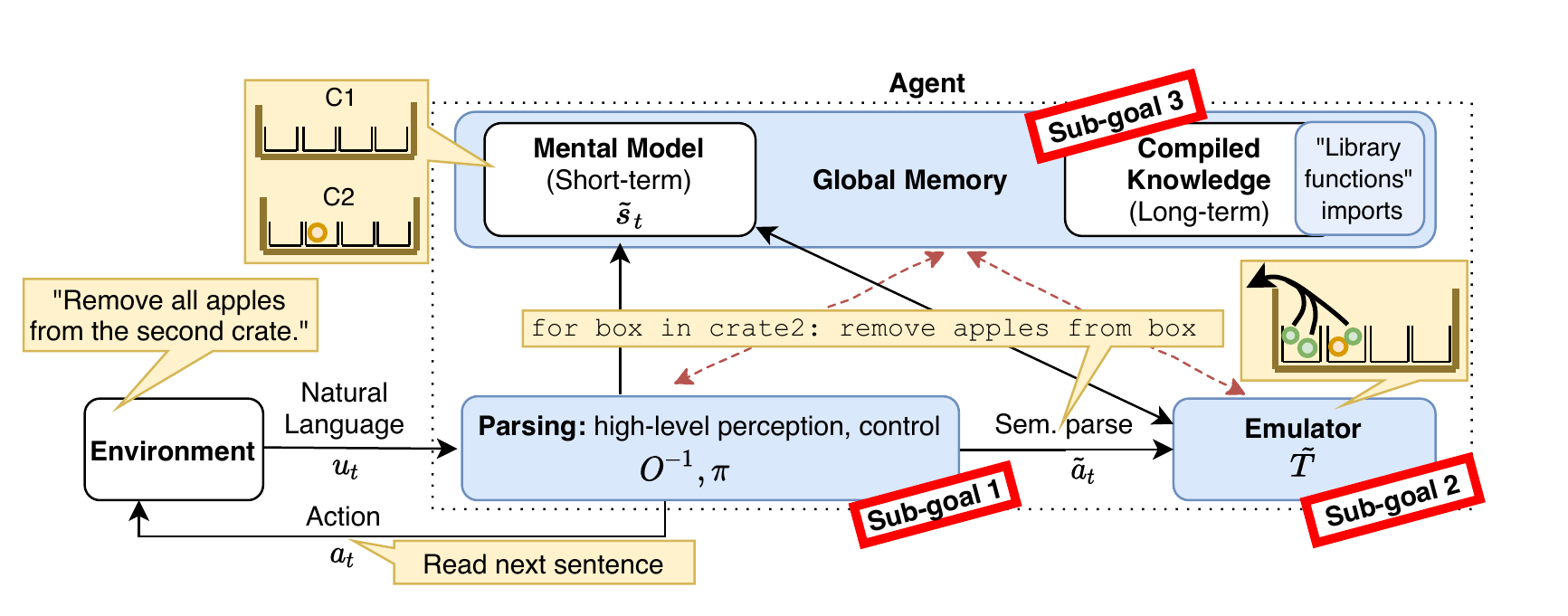}
\caption{\label{fig:ecog-lm} Architecture for comprehender (\S\ref{sec:ecl-language-model}), demonstrated on a symbolic version of the example task of Fig. \ref{fig:scone-in-a-box}. The agent receives natural language input from the environment. The agent has global memory -- short-term, keeping track of the mental model of the world, and long-term, containing compiled knowledge (``library classes and functions''). The parser interprets input to parse $\intaction_{t}$ enacting mental simulation using emulator. The mental model is then updated, ready for the next input. The sub-goals refer to the order in which components are learned (as opposed to hard-coded) in our proposed roadmap (\S\ref{sec:roadmap}).}
\end{figure*}

This section presents a schematic ECL-inspired architecture towards the implementation of the \textbf{comprehension model} ($C$), which addresses the challenges presented in \S\ref{ssec:emb-challenges}.
Fig. \ref{fig:ecog-lm} shows the proposed architecture. For simplicity, the focus is on a static reading comprehension setting, but the architecture supports richer environments as well.

\subsection{Environment}
The environment provides an ``interaction API'' to the agent, as well as the reward signal. The supported interaction may vary considerably depending on the task; for reading comprehension, it {allows structured access to the text} while supporting flexible reading strategies~\citep{yuan2019interactive}. The flexibility is important for long documents, where navigation may be required~\citep{geva-berant-2018-learning}. For executable semantic parsing, there might be external systems to interact with besides the text, such as a database~\citep{Liang2016}. 

\subsection{Agent}
The agent architecture approximates the important ECL functions outlined in \S\ref{sec:ecl-link}, and consists of four main modules:

\xhdr{\textit{Memory}} 
We distinguish between two forms of memory, the first an episodic, short-term \textbf{mental model} -- the system's current abstract state representation ($\astate_{t}$). The symbolic programming analog is the execution trace of a program, containing the states of relevant working variables at each execution step. Fig. \ref{fig:ecog-lm} displays the updated mental model, after the removal of the apples. \textbf{Compiled knowledge}, or long-term memory, reflects highly familiar object representations, behaviors and schemata, such as common sense, intuitive psychology and physics. The symbolic programming language analogs of this are libraries; largely static, hierarchical and compositional repositories of functions and classes. In the course of language interpretation, these libraries are ``importable'': for the symbolic example in Fig. \ref{fig:ecog-lm}, the parser might instantiate a new variable of an imported type (e.g., \texttt{crate2 = Container()}). Both types of memory are {accessible} for all components of the agent.

\xhdr{\textit{Parser}} Abstraction of higher-level perception, control, reasoning and linguistic functions. Handles interpretation of new linguistic inputs based on prior knowledge and the current mental state. Consonant with the view of analogy-making as a kind of higher-level perception or recognition~\citep{MitchellAnalogy1993}, \textbf{metaphoric inference} is involved in grounding a novel input $u_{t}$ into \textbf{internal, neural} state representations $\intstate_{t}$, $\intaction_{t}$ affording simulation. See Fig. \ref{fig:ecog-lm} and Fig. \ref{fig:metalang-challenge} for examples on literal and non-literal language, respectively.

\xhdr{\textit{Emulator}} Functionally similar to the executor module in executable semantic parsing, but learned, and obviously far greater in scale. This module is an abstraction of neural emulation mechanisms ($\Temu$), representing a wide range of functions, from lower-level motor control and imagery to higher-level models used for planning and theory of mind~\citep{Grush2004}. It operates over the current mental model and semantic parse from the parser. The output is then an updated mental model.
\paragraph{}

Importantly, the proposed architecture is designed to address the challenges outlined in \S\ref{ssec:emb-challenges}; compiled knowledge underlies \textbf{human common ground}, the building blocks of $\intstate$, $\intaction$ and $\Temu$. Memory and emulation are instrumental for \textbf{accumulation in discourse}. The ability to understand \textbf{abstract language} involves all modules in the system.

\section{Implementation Roadmap}
\label{sec:roadmap}
The architecture outlined in \S\ref{sec:emb-arch} is very ambitious; its implementation requires much further research. This section proposes a \emph{roadmap} to this goal, identifying three sub-goals (Fig. \ref{fig:ecog-lm}), presented in order of increasing difficulty. Broadly speaking, the level of difficulty is determined by which components are assumed as \emph{given} in the input (here this also means they are hard-coded in a symbolic programming language), and which must be \emph{learned}.

\subsection{Sub-goal 1: learning open-domain simulation}
\label{ssec:phase-one}

Observing that literal language is close to the embodied primitives level, its interpretation is simpler (than that of non-literal language, see \S\ref{sec:ecl-link}). Therefore, in this phase, the emulator and compiled knowledge are hard-coded; here the focus is learning the parser. In other words, this sub-goal focuses on extending executable semantic parsing from relatively narrow domains to handle more general \emph{literal language} on-the-fly, similarly to zero-shot semantic parsing~\citep{Givoli2019}. 

For the example in \S\ref{ssec:ootb-reason}, the parser could be expected to infer the types (boxes as containers, fruits as objects) either by context (\citet{Yao2018} explore a preliminary schema-based approach) or explicit declarative language, using them to \textbf{configure the emulator} to handle the specific required problem setting~\citep{tamari2020ecological}.

As in similar projects exploring embodied understanding~\citep{pustejovsky2016voxml,Baldridge2018}, new simulator frameworks must be developed. %
While full embodiment calls for multiple modalities, the degree to which it is required remains an important open question~\citep{Lupyan2019}. Accordingly, and for immediate applicability to purely textual NLU problems we propose also focusing on the simpler setting of interactive text~\citep{Nelson2005}. Recent research on text-based games shows how agents can learn to ``program'' in such languages~\citep{Cote,ammanabrolu-riedl-2019-playing}, and how real language understanding problems can be framed as executable semantic parsing using configurable text-based simulators~\citep{tamari-etal-2019-playing}.

\subsection{Sub-goal 2: learning to simulate}
\label{ssec:phase-two}

This phase assumes that the compiled knowledge is given (hard-coded), and the parsing and emulator modules are neural (learned). A hard-coded emulator will likely be needed to train a learned emulator. The learned event execution of \citet{narayanan1997knowledge} provides a useful starting point towards computational models capable of such inference. In general, learned simulation is relatively unexplored in the context of natural language, though recent work has explored it in  generated instruction following setups~\citep{gaddy-klein-2019-pre,adhikari2020learning}.  Outside of NLU, learning structured world models is a long-studied, fast-growing field in embodied AI research~\citep{Schmidhuber1990MakingTW, Ha2018,Hamrick2019, Kipf2020Contrastive}, and recently also in learned executors for neural programming~\citep{Kant2018}. We expect much useful cross fertilization with these fields.

\subsection{Sub-goal 3: learning compiled knowledge}
\label{ssec:phase-three}
This phase focuses on the component seemingly hardest to learn -- compiled knowledge. Out of scope here is fully neural setting where all components are jointly learned, as in continual learning research~\citep{parisi2019continual}. Instead, we focus on a simpler setting, in which the compiled knowledge is learned but represented by symbolic code; i.e., learning the static code library underlying the simulation framework. This sub-goal is relevant for training the parser (\S\ref{ssec:phase-one}) as well as the emulator (\S\ref{ssec:phase-two}), and can be pursued in parallel to them.

In this setting, learning compiled knowledge is closely related to never-ending language learning~\citep{mitchell2018never}, interactive task learning~\citep{she-chai-2016-incremental}, automated knowledge base construction~\cite{winn2019alexandria}, and frame induction from text~\citep{qasemizadeh-etal-2019-semeval}. Our proposed paradigm suggests enriching classic symbolic knowledge representations~\citep{speer2017conceptnet} to executable form~\citep{tamari2020ecological}. Preliminary steps in this direction are seen in inferential knowledge bases such as ATOMIC~\citep{sap2019atomic}, which provides limited execution logic using edges typed with if-then relations.

Alongside FrameNet and MetaNet, others have collected schema and metaphor mappings, by learning them from large corpora~\citep{beigman-klebanov-etal-2016-semantic,gao2018neural}. \citet{pastra2011embodied} built a database of concepts directly groundable to sensorimotor representations, primarily for robotics applications.

\section{Conclusions}
\label{sec:conclusions}
This position paper has proposed an approach to representation and learning based on the tenets of ECL. The proposed architecture, drawing on contemporary cognitive science, aims to address key limitations of current NLU systems through mental simulation and grounded metaphoric inference. We outlined major challenges and suggested a roadmap towards realizing the proposed vision.

Growing empirical evidence shows that language is intricately intertwined with a vast range of other neural processes. Accordingly, this work suggests a symbiotic view of cognitive science, embodied AI, and computational linguistics. By sharing common foundational problems, these fields may better share and co-evolve common solutions.
Finally, we believe that attaining deeper language understanding must be a large scale effort, beyond the scope of any one research group. We hope that the paradigm presented here will help provide coherence to such efforts. One of our main goals was to stimulate a discussion; moving forward, we welcome comments, feedback, and suggestions. %

\section*{Acknowledgments} 
We thank the reviewers for their insightful comments. We further thank Ari Rappoport, the Hyadata Lab at HUJI, Yoav Goldberg, Ido Dagan, Jonathan Berant, and the BIU \& TAU NLP seminar audiences for interesting discussion and thoughtful remarks.
This work was supported by the European Research Council (ERC) under the European Union's Horizon 2020 research and innovation programme (grant no. 852686, SIAM) and NSF-BSF grant no. 2017741 (Shahaf), as well as the Israel Science Foundation grant no. 929/17 (Abend).

\bibliography{refs,anthology}
\bibliographystyle{acl_natbib}

\appendix

\end{document}